\documentclass[lettersize,journal]{IEEEtran}

\usepackage{amsmath,amsfonts}
\usepackage{array}
\usepackage[caption=false,font=normalsize,labelfont=sf,textfont=sf]{subfig}
\usepackage{textcomp}
\usepackage{stfloats}
\usepackage{url}
\usepackage{verbatim}
\usepackage{graphicx}
\usepackage{cite}

\usepackage{booktabs}
\usepackage{enumitem}  %
\usepackage{colortbl}  %
\usepackage{xcolor}
\usepackage{multirow}
\usepackage[ruled]{algorithm2e}
\makeatletter
\newcommand{\removelatexerror}{\let\@latex@error\@gobble}
\makeatother

\begin{document}

\title{Towards Video Anomaly Retrieval from Video Anomaly Detection: New Benchmarks and Model}

\author{Peng Wu, Jing Liu~\IEEEmembership{Senior Member, IEEE}, Xiangteng He, Yuxin Peng~\IEEEmembership{Senior Member, IEEE},\\ Peng Wang, and Yanning Zhang~\IEEEmembership{Senior Member, IEEE}
\thanks{Peng Wu, Peng Wang, and Yanning Zhang are with the National Engineering Laboratory for Integrated Aero-Space-Ground-Ocean Big Data Application Technology, School of Computer Science, Northwestern Polytechnical University, China. E-mail: xdwupeng@gmail.com; {peng.wang, ynzhang}@nwpu.edu.cn.
Jing Liu is with the Guangzhou Institute of Technology, Xidian University, China. E-mail: neouma@163.com.
Xiangteng He and Yuxin Peng are with the Wangxuan Institute of Computer Technology, Peking University, China. E-mail: {hexiangteng, pengyuxin}@pku.edu.cn. This work is supported by the National Natural Science Foundation of China (No. 62306240, U23B2013, U19B2037, 62272013, 61925201, 62132001), China Postdoctoral Science Foundation (No. 2023TQ0272), and the Fundamental Research Funds for the Central Universities (No. D5000220431). (Corresponding author: Yanning Zhang.)}

\thanks{Manuscript received April 19, 2021; revised August 16, 2021.}}

\markboth{Journal of \LaTeX\ Class Files,~Vol.~14, No.~8, August~2021}%
{Shell \MakeLowercase{\textit{et al.}}: A Sample Article Using IEEEtran.cls for IEEE Journals}


\maketitle

\begin{abstract}
Video anomaly detection (VAD) has been paid increasing attention due to its potential applications, its current dominant tasks focus on online detecting anomalies
, which can be roughly interpreted as the binary or multiple event classification. However, such a setup that builds relationships between complicated anomalous events and single labels, e.g., ``vandalism'', is superficial, since single labels are deficient to characterize anomalous events. In reality, users tend to search a specific video rather than a series of approximate videos. Therefore, retrieving anomalous events using detailed descriptions is practical and positive but few researches focus on this. In this context, we propose a novel task called \textbf{V}ideo \textbf{A}nomaly \textbf{R}etrieval (\textbf{VAR}), which aims to pragmatically retrieve relevant anomalous videos by cross-modalities, e.g., language descriptions and synchronous audios. Unlike the current video retrieval where videos are assumed to be temporally well-trimmed with short duration, VAR is devised to retrieve long untrimmed videos which may be partially relevant to the given query. To achieve this, we present two large-scale VAR benchmarks and design a model called \textbf{A}nomaly-\textbf{L}ed \textbf{A}lignment \textbf{N}etwork (\textbf{ALAN}) for VAR. In ALAN, we propose an anomaly-led sampling to focus on key segments in long untrimmed videos. Then, we introduce an efficient pretext task to enhance semantic associations between video-text fine-grained representations. Besides, we leverage two complementary alignments to further match cross-modal contents. Experimental results on two benchmarks reveal the challenges of VAR task and also demonstrate the advantages of our tailored method. Captions are publicly released at https://github.com/Roc-Ng/VAR.
\end{abstract}


\begin{figure}[t]
  \centering
   \includegraphics[width=1.0\linewidth]{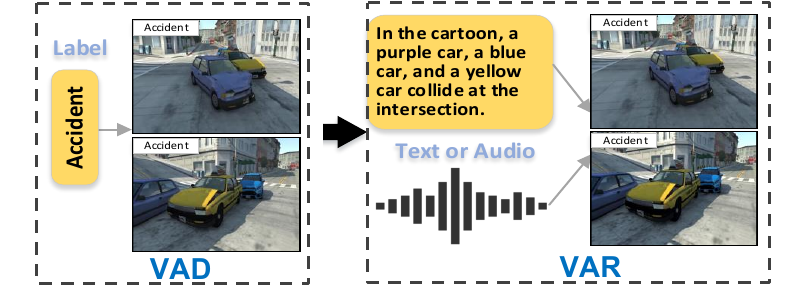}
   \caption{VAD \textit{vs.} VAR. Single labels may be unable to describe sequential anomalous events in VAD, but text captions or synchronous audios can sufficiently depict events in VAR.}
   \label{fig:varvad}
\end{figure}

\section{Introduction}
\label{sec:intro}
\IEEEPARstart{V}{ideo} anomaly detection (VAD) plays a critical role in video content analysis, and has become a hot topic being studied due to its potential applications, e.g., danger early-warning. VAD, by definition, aims to identify the location of anomaly occurrence, which can be regarded as the frame-level event classification. VAD can be broadly divided into two categories, i.e., semi-supervised~\cite{sabokrou2017deep, liu2018future, wu2019deep, park2020learning, georgescu2021anomaly} and weakly supervised~\cite{sultani2018real, wu2020not, feng2021mist,tian2021weakly,wu2021weakly}. The former typically recognizes anomalies through self-supervised learning or one-class learning. The latter, thanks to massive normal and abnormal videos with video-level labels, achieves better detection accuracy.

Impressive progress has been witnessed for VAD, however, an event in videos generally captures an interaction between actions and entities that evolves over time, simply utilizing single labels in VAD may be insufficient to explain the sequential events depicted. Besides, compared with VAD, offline video search thus far is still more commonly used in real-world applications. Imagining the case when searching for related videos, we prefer to use comprehensive descriptions to accurately search, e.g., ``At night, two topless men smashed the door of the store.'', rather than use a single coarse word, e.g, ``vandalism'', to get a large collection of rough results.

Based on VAD, we propose a new task called Video Anomaly Retrieval (VAR) and present two large-scale benchmark datasets, UCFCrime-AR and XDViolence-AR, to further facilitate the research of video anomaly analysis. The goal of VAR is to retrieve relevant untrimmed videos given cross-modal queries, e.g., text captions and synchronous audios, and vice versa. Unlike VAD, VAR depicts anomalies from multiple viewpoints and sufficiently characterizes sequential events. We illustrate the advantage of video anomaly retrieval in Figure~\ref{fig:varvad}. {VAR task has high value to real-world applications, especially for smart ground and car surveillance. Generally speaking, for surveillance, the recorded video will be stored in the hard disk or memory card as a series of segments with a certain time length. After an abnormal event occurs, we need to search the corresponding video segment that contains the queried abnormal event through the descriptions, such as a white car crashed into the rear of a van, a group of people breaking into a house at night, etc.}

\begin{figure}[t]
  \centering
   \includegraphics[width=1.0\linewidth]{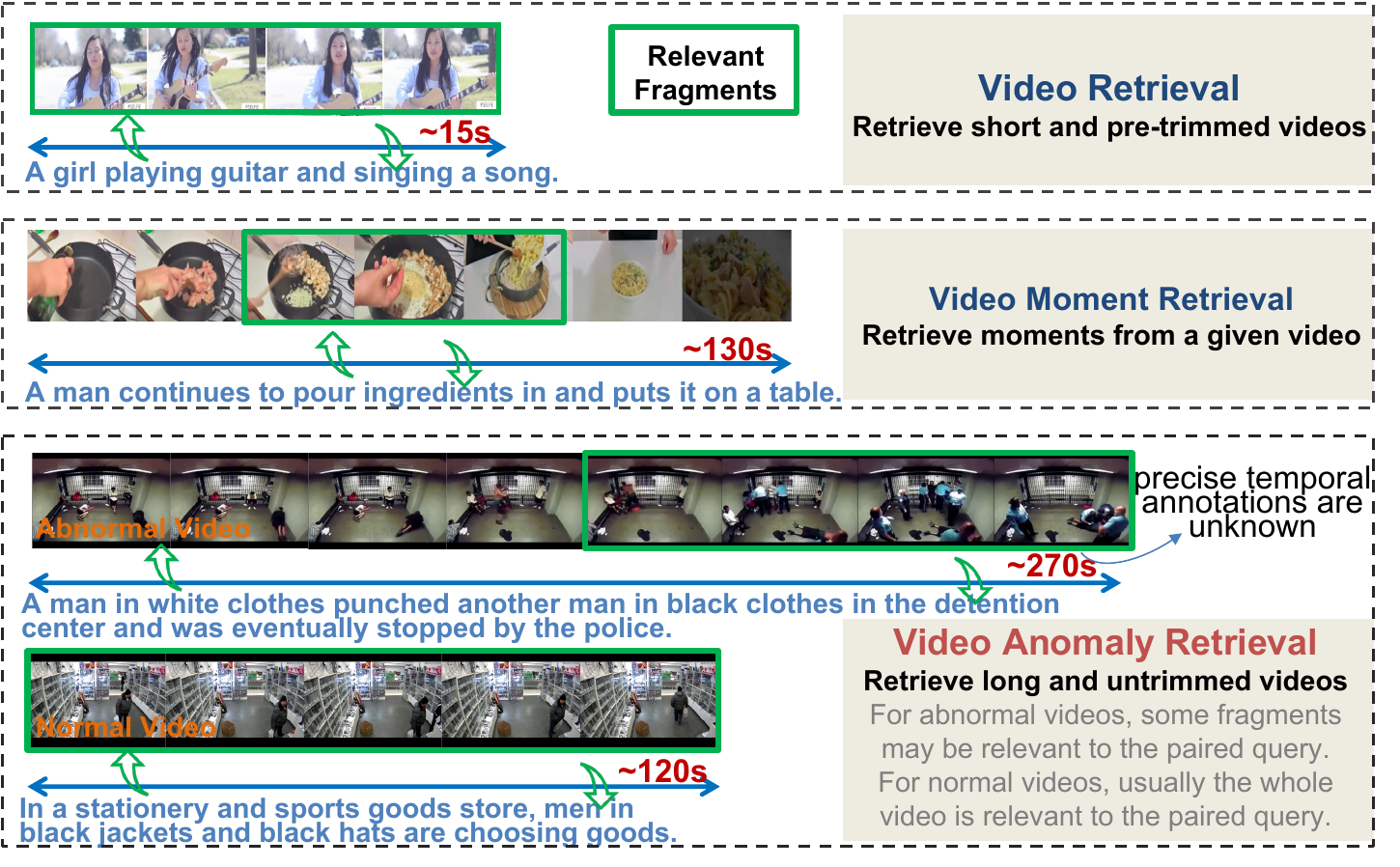}
   \caption{Comparison of VAR with video retrieval and video moment retrieval.}
   \label{fig:VARcom}
 
\end{figure}

Our VAR is considerably different from traditional video retrieval (VR)~\cite{miech2018learning, liu2019use, gabeur2020multi}. In traditional video retrieval, videos are assumed to be temporally pre-trimmed with short duration, and thus the whole video is supposed to be completely relevant to the paired query. In reality, videos are usually not well-trimmed and may only exist partial fragments to fully meet the query. In VAR, the main goal is to retrieve \textit{long} and \textit{untrimmed} videos. Such a setup more meets realistic requirements, and also evokes the new challenge. Concretely, the length of relevant fragments is variable in videos w.r.t. given paired queries. For normal videos, the relevant fragment is generally the whole video; For abnormal videos, the relevant fragment may occupy only a fraction or a lion's share of the entire video since the length of anomalous events is inconstant in videos. Besides, our VAR task also differs from video moment retrieval (VMR)~\cite{yang2022video, yang2022tubedetr,cui2022video, wang2022cross} since the latter is to retrieve moments rather than untrimmed videos. Because both abnormal videos and normal videos (no anomaly) need to be retrieved in VAR, video moment retrieval methods are hard to tackle this task. Traditional video retrieval and video moment retrieval methods cannot solve this new challenge well, detailed results are listed in Tables~\ref{tab:comparison_ucf} and~\ref{tab:comparison_xd}. The differences between video retrieval, video moment retrieval and video anomaly retrieval are shown in Figure~\ref{fig:VARcom}.

To overcome the above challenge, we propose \textbf{ALAN}, an \textbf{A}nomaly-\textbf{L}ed \textbf{A}lignment \textbf{N}etwork for video anomaly retrieval. 
In ALAN, video, audio, and text encoders are intended to encode raw data into high-level representations, and cross-modal alignment is introduced to match cross-modal representations from different perspectives. Since videos are long untrimmed and anomalous events have complex variations in the scenario and length, we expect that, in the encoding phase, the retrieval system maintains holistic views, meanwhile, focuses on key anomalous segments, so that cross-modal representations can be well aligned in the joint embedding space. Therefore, vanilla fixed-frame sampling, e.g., uniform sampling and random sampling, is not flexible to focus on specific anomalous segments. Inspired by dynamic neural networks~\cite{han2021dynamic, rao2021dynamicvit, zhi2021mgsampler,fayyaz2022adaptive}, we propose an anomaly-led sampling, which simply resorts to frame-level anomaly priors generated by an ad-hoc anomaly detector and does not require intensive pairwise interactions between cross modality, to select key segments with large anomaly identification degree. 
We then couple these two win-win sampling mechanisms for videos and audios, where anomaly-led sampling focuses on anomalous segments, and fixed-frame sampling pays attention to the entirety as well as normal videos. Furthermore, to establish associations between video-text fine-grained representations as well as maintain high retrieval efficiency, we also propose a pretext task, i.e., video prompt based masked phrase modeling (VPMPM), serving the model training. Particularly, a new module termed Prompting Decoder takes both frame-level video representations and contextual text representations as input and predicts the masked noun phrases or verb phrases by the cross-modal attention, where video representations serve as fixed prompts~\cite{li2021align,hou2022milan}. In this paper, video frames are regarded as the fine granularity as frames usually reflect more detailed content of videos, meanwhile noun phrases and verb phrases in texts, e.g., ``a black car'' and ``left quickly'', are regarded as the fine granularity, which reflect the local spatial contents and temporal dynamics in the video, respectively. Notably, compared with nouns and verbs, noun phrases and verb phrases contain more contents, and can also particularly illustrate the subtle differences. Finally, such a proxy training objective optimizes the encoder parameters and further promotes the semantic associations between local video frames and text phrases by cross-modal interactions.

To summarize, our contributions are three-fold:
\begin{itemize}
\item
We introduce a new task named video anomaly retrieval to bridge the gap between the literature and real-world applications in terms of video anomaly analysis. To our knowledge, this is the first work moves towards VAR from VAD;
\item
We present two large-scale benchmarks, i.e., UCFCrime-AR and XDViolence-AR, based on public VAD datasets. The former is applied to video-text VAR, the latter is to video-audio VAR;
\item
We propose a model called ALAN, aiming at challenges in VAR, where anomaly-led sampling, video prompt based masked phrase modeling, and cross-modal alignment are introduced for the attention of anomalous segments, enhancement of fine-grained associations, and multi-perspective match, respectively. 
\end{itemize}

\section{Related Work}
\subsection{Video Anomaly Detection}
Aided by the success of deep learning, VAD has made much good progress in recent years, which is usually classified into two groups, semi-supervised anomaly detection and weakly supervised anomaly detection. In semi-supervised anomaly detection, only normal event samples are available for model training. Recent researchers mainly adopt deep auto-encoders~\cite{liu2018future,lee2019bman, ionescu2019object,gong2019memorizing,park2020learning, wang2022video,yang2022dynamic, yang2023video, yan2023feature} for self-supervised learning, e.g., reconstruction, prediction, jigsaw, etc. Weakly supervised anomaly detection can be regarded as the problem of binary classification, to obtain frame-level predictions given coarse labels, and multiple instance learning~\cite{sultani2018real,lv2021localizing, wu2021learning,feng2021mist,tian2021weakly, cao2022adaptive, huang2022weakly} is widely used to train models. Unlike VAD that utilizes single labels to distinguish whether each frame is anomalous or not, our proposed VAR uses elaborate text descriptions or synchronous audios to depict the sequential events.

\subsection{Cross-Modal Retrieval}
We mainly introduce cross-modal retrieval~\cite{peng2017overview, peng2017cross, yu2022coca,zuo2023fine} built on videos, texts, and audios. There are some works~\cite{monfort2021spoken,oncescu2021audio, gabeur2021masking, rouditchenko2021cascaded,morgado2021audio,shen2023end} focus on audio-text and audio-video retrieval. {Specifically, Tian et al.~\cite{tian2018audio} propose an audio-to-video/video-to-audio cross-modality localization/retrieval task~\cite{wei2022learning}, i.e, given a sound segment, locate the corresponding visual sound source temporally within a video, and vice versa. Then Wu et al.~\cite{wu2019dual} introduce a novel dual attention matching method for this task. Recently, Lin et al.~\cite{lin2023vision} propose a latent audio-visual hybrid adapter that adapts pre-trained vision transformers to audio-visual tasks, this method focuses on audio-video event localization task rather than cross-modal retrieval.} In addition, text-video retrieval is a key role in cross-modal retrieval. Generally, text-video retrieval can be divided into two categories, i.e., dual-encoder and joint-encoder. Dual-encoder based methods usually train two individual encoders to learn video and text representations and then align these representations in joint embedding spaces. Among them, some works~\cite{li2019w2vv++, gabeur2020multi,dong2021dual, liu2021hit} focus on learning single global representations, but they lack the consideration of fine-grained information. Thereby, several works devote efforts to aligning fine-grained information~\cite{wray2019fine, wu2021hanet, han2021fine, yang2021taco, wangdig,ge2022bridging,ma2022x}. Joint-encoder based methods~\cite{li2020hero,lei2021less,luo2020univl,ji2022cret} typically feed the video and text into a joint encoder to capture their cross-modal interactions. In comparison to dual-encoder based methods, joint-encoder based methods explicitly learn fine-grained associations and achieve more impressive results, but sacrifice the retrieval efficiency since every text-video pair needs to be fed into the encoder at the inference time.

Different from the above video retrieval, we consider a more realistic scenario, where most videos contain anomalous events, and a more realistic demand, where videos are long untrimmed and partially relevant to the cross-modal query~\cite{dong2022partially}. Such a new task poses extra challenges as well as multi-field research points. In addition, our ALAN also differs from video moment retrieval methods~\cite{gao2017tall,zhang2019man,mithun2019weakly,ding2022exploring} in that it does not require complex cross-modal interactions.

\section{Benchmark}
Manually collecting a large-scale video benchmark is labor-intensive and time-consuming, it is also subjective since video understanding can often be an ill-defined task with low annotator consistency~\cite{miech2019howto100m}. Therefore, we start with two acknowledged datasets in VAD community, i.e., UCF-Crime~\cite{sultani2018real} and XD-Violence~\cite{wu2020not}, and construct our benchmarks for VAR. We adopt these two datasets as the base since they thus far are the two most comprehensive VAD datasets in terms of length and scene, where the total duration of them are 128 and 217 hours, respectively. Besides, they are also collected from a variety of scenarios. For example, UCF-Crime covers 13 real-world anomalies as well as normal activities, and XD-Violence captures 6 anomalies and normal activities from movies and YouTube. In addition, both of them contain half normal videos and half abnormal videos, therefore, retrieval systems retrieve both abnormal and normal videos from the video gallery given related cross-modal queries in VAR.
Large and diverse video databases allow us to construct more practicable benchmarks for VAR.

\subsection{UCFCrime-AR}
UCF-Crime dataset consists of 1900 untrimmed videos with 950 abnormal videos and 950 normal videos. Notably, for anomaly videos in the training set, the start timestamp and duration of anomalous activities are unavailable. For normal videos, they are totally anomaly-free. We directly use the total videos as the video search base. To achieve cross-modal retrieval, we require pairwise text descriptions.

We invite 8 experienced annotators who are proficient in Chinese and English to annotate these videos. The annotators watch the entire video and make the corresponding captions in both Chinese and English. Specifically, annotators are required to focus on anomalous events when describing anomaly videos. Due to the subtle differences in videos for the same anomaly category, we need to obtain quality sentence annotations to distinguish fine differences and avoid being into a \textit{one-to-many} dilemma~\cite{wray2021semantic} which often appears in the current video retrieval. To be specific, there are at most two annotators to describe videos in the same category. For two similar videos in the same category, describe their differences in detail as much as possible. Take the scene of a fighting between two people as an example, e.g., ``At a party, the yellow-haired man suddenly attacked a man opposite him.'', ``A young man suddenly beat another man with glasses in the elevator.'' The above two annotations clearly describe the difference between two similar videos. Finally, we double-check each sentence description to guarantee the quality.

Following the partition of UCF-Crime, UCFCrime-AR includes 1610 training videos and 290 test videos. Each video is annotated with captions in both English and Chinese. In this work, we only use captions in English.

\subsection{XDViolence-AR}
As for XD-Violence, we found that it is very hard to describe videos in a few sentences due to their complicated contents and scenarios. Hence we changed focus and started a new line of audio-to-video retrieval due to its natural audio-visual information, that is, we use videos and synchronous audios for cross-modal anomaly retrieval. Unlike texts, audios have the same granularity as videos. Similar to UCF-Crime, XD-Violence is also a weakly supervised dataset, namely, frame-level annotations are unknown. XDViolence-AR is split into two subsets, with 3954 long videos for training and 800 for testing.

\subsection{Benchmark Statistics}

\begin{figure}[!tb]
  \centering
   \includegraphics[width=0.9\linewidth]{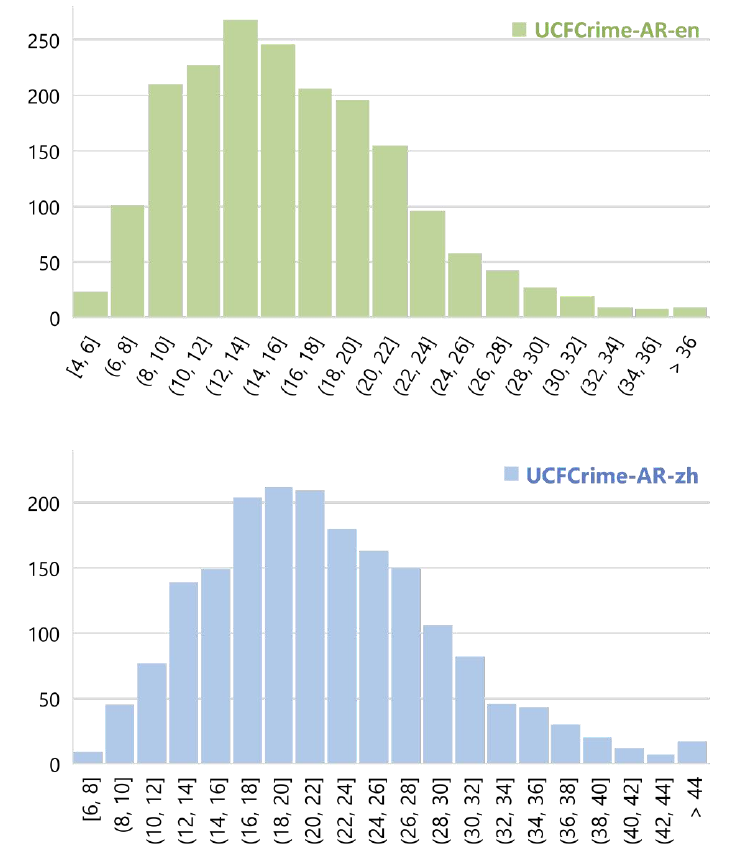}
   \caption{Statistical histogram distributions on UCFCrime-AR. Left: text captions in English; Right: text captions in Chinese.}
   \label{fig:data}
\end{figure}

We compare two benchmarks with several cross-modal retrieval/location datasets in Table~\ref{tab:describe}. As we can see that, video databases in UCFCrime-AR and XDViolence-AR are both large-scale and are made public in recent years, where the former is applied to video-text (V-T) anomaly retrieval, and the latter is applied to video-audio anomaly retrieval (V-A). Notably, the average length of videos in VAR benchmarks is significantly longer than that of videos in traditional video retrieval datasets. For example, the average length of videos of UCFCrime-AR and XDViolence-AR are 242s and 164s, whereas that of MSR-VTT~\cite{xu2016msr}, VATEX~\cite{wang2019vatex}, {VAE~\cite{tian2018audio}}, AudioCaps~\cite{kim2019audiocaps}, and {LLP~\cite{tian2020unified}} are in the range of 10s to 20s, {and TVR~\cite{lei2020tvr} is mainly applied to video moment retrieval task, its average length of videos is still much shorter than our benchmarks.} Longer videos emphasize again the goal of VAR is to retrieve long and untrimmed videos, such a setup meets realistic requirements, and also reveals VAR is a more challenging task. For video-text UCFCrime-AR, we also present histogram distributions of captions in Figure~\ref{fig:data}. The average caption lengths of UCFCrime-AR-en, UCFCrime-AR-zh are 16.3 and 22.4, which are longer than those of previous datasets in video retrieval. e.g., the average caption lengths of VATEX-en~\cite{wang2019vatex}, VATEX-zh~\cite{wang2019vatex}, and MSR-VTT~\cite{xu2016msr} are 15.23, 13.95, and 9.28, respectively.

\begin{table}[t]
  \centering
  \caption{Comparison of UCFCrime-AR and XDViolence-AR with several video-text and video-audio retrieval/localization datasets.}
  \resizebox{1\columnwidth}{!}{
  \begin{tabular}{l|rrrcc}
    \toprule
     Datasets & Duration & \#Videos & Avg.len. & Type & Year\\
    \hline
     MSR-VTT~\cite{xu2016msr} & 40h  & 7.2k & 20s &  V-T & 2016\\
     VATEX~\cite{wang2019vatex} & 114h & 41k & 10s & V-T & 2019\\
      {TVR~\cite{lei2020tvr}}&{463h}&{21.8k}&{76s}&{V-T}&{2020}\\
     {AVE~\cite{tian2018audio}}&{11h}&{4.1k}&{10s}&{V-A}&{2018}\\
     AudioCaps~\cite{kim2019audiocaps} & 127h & 46k & 10s & V-A & 2019\\
     {LLP~\cite{tian2020unified}}&{33h}&{11k}&{10s}&{V-A}&{2020}\\
     \hline
     UCFCrime-AR & 128h  & 1.9k & \textbf{242s} &  V-T & 2018\\
     XDViolence-AR & 217h & 4.8k & \textbf{164s} & V-A & 2020 \\
    \bottomrule
  \end{tabular}}
   \label{tab:describe}
\end{table}

\section{Method}
In this section, we introduce ALAN in detail. In Sec.~\ref{sec: encoder}, we first introduce three encoders in ALAN, namely, video encoder, text encoder, and audio encoder, the goal of these encoders is to project raw videos, texts, and audios into high-level representations. In Sec.~\ref{sec: anomaly-led sampling}, we introduce the anomaly-led sampling mechanism which is utilized in both video encoder and audio encoder. In Sec.~\ref{sec: mpm}, we describe a novel pretext task, i.e., VPMPM, which is applied to video-text anomaly retrieval. At last, we describe the cross-modal alignment and training objectives in Secs.~\ref{sec: alignment} and~\ref{sec: training objectives}.

\begin{figure*}[t]
  \centering
   \includegraphics[width=1.0\linewidth]{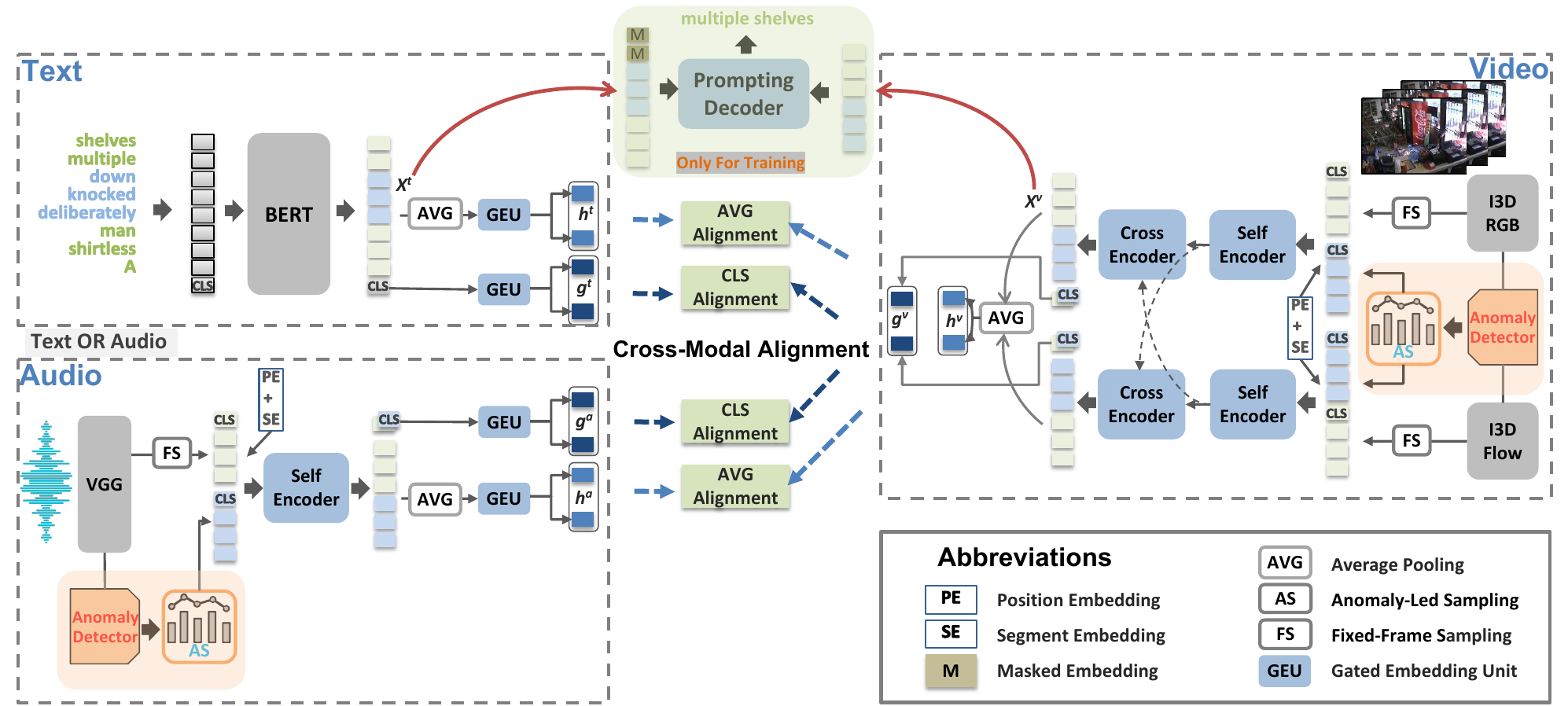}
   \caption{Overview of our ALAN. It consists of several components, i.e., video encoder, text encoder, audio encoder, pretext task VPMPM, and cross-modal alignment.}
   \label{fig:pipeline}
\end{figure*}

\subsection{Encoders}
\label{sec: encoder}
\noindent \textbf{Video encoder.} Unlike images, videos possess space-time information~\cite{arnab2021vivit, neimark2021video}. As a consequence, we consider both appearance and motion information to encode videos. Specifically, given an video $v$, we use I3D-RGB and I3D-Flow pre-trained on Kinetics~\cite{carreira2017quo} to extract frame-level object and motion features, respectively, then project these features into a $d$-dimensional space for the subsequent operations. Here, object and motion feature sequences are denoted as $\boldsymbol{F}^o(v)$ and $\boldsymbol{F}^m(v)$, respectively. Both sequences contain $T$ clip features. For the sake of clarity, we use $\boldsymbol{F}(v)$ to denote $\boldsymbol{F}^o(v)$ and $\boldsymbol{F}^m(v)$. Taking into account the variety of anomalous event duration in untrimmed videos, we sample two sparse video clips with different concerns, i.e., $\boldsymbol{U}$ and $\boldsymbol{R}$, from $\boldsymbol{F}(v)$ by means of the fixed-frame sampling and our proposed anomaly-led sampling.

As demonstrated in Figure~\ref{fig:pipeline}, the video encoder is a symmetric two-stream model, one stream takes as input object, and the other takes as input motion. {In order to fuse features in different modalities and different temporalities for final representations, we employ the Transformer~\cite{vaswani2017attention} as the base model, which has been widely used in VAD and VR tasks with good results. For example, Huang et al.~\cite{huang2022weakly, huang2022hierarchical} and Zhao et al.~\cite{zhao2022centerclip} used Transformer to tackle VAD and VR tasks, respectively}. We first concatenate two different sampling clips as a new sequence, i.e., $[U_{CLS}, U_1, ..., R_N, R_{CLS}, R_1, ..., R_N]$, where $U_{CLS}$ and $R_{CLS}$ are [CLS] tokens, which are the average aggregation of all features in $\boldsymbol{U}$ and $\boldsymbol{R}$, respectively. Then, we add positional embeddings~\cite{vaswani2017attention} and sequence embeddings to this sequence. Here positional embeddings provide temporal information about the time in the video, and sequence embeddings depict that features in $\boldsymbol{U}$ and $\boldsymbol{R}$ stem from different sequences. In video encoder, Self Encoder is devised to capture contextual information, which is a standard encoder layer in Transformer. The following Cross Encoder takes the self-modality as the query, and cross-modality contextual features as the key and value to encode cross-modal representations through cross-modal attention. Cross Encoder is composed of multi-head attention, a linear layer, a residual connection, and a layer normalization. Finally, we obtain two different video representations, one is the average of output from $U_{CLS}$ and $R_{CLS}$, denoted as $g^v$ (including $g^{vo}$ and $g^{vm}$), the other is the mean of average pooling aggregation of output from $\boldsymbol{U}$ and $\boldsymbol{R}$, denoted as $h^v$ (including $h^{vo}$ and $h^{vm}$). Such a simple pooling operation is parameter-free and effective in our work, enabling $h^v$ to involve local fine-grained information.

\noindent \textbf{Text encoder.} Give a text caption $t$, we aim to learn the alignment between it and a related video at two different levels. At first, we leverage a pre-trained BERT~\cite{gabeur2020multi} to extract features aided by its widespread adoption and proven performance in language representations. Following the video encoder, we obtain $g^t$ from the [CLS] output of BERT, and $h^t$ by using the average pooling operation for word-level representations. To match the object and motion representations of videos, here we use the gated embedding unit~\cite{miech2018learning} acting on $g^t$ and $h^t$ to produce $g^{to}$, $g^{tm}$ and $h^{to}$, $h^{tm}$, respectively.

\noindent \textbf{Audio encoder.} Give an audio $a$, we first extract audio features using a pre-trained VGGish~\cite{gemmeke2017audio}, and project these features into the $d$-dimensional space. As shown in Figure~\ref{fig:pipeline}, the audio encoder is similar to video encoder in terms of structure. The difference lies in that audio encoder is a single-stream model and has no Cross Encoder. In a similar vein, two different audio representations $g^a$ and $h^a$ are obtained. The gated embedding unit is also applied to match the object and motion representations of videos.

\subsection{Anomaly-Led Sampling}
\label{sec: anomaly-led sampling}

As mentioned, only fixed-frame sampling (FS) cannot capture variable anomalous events in anomaly videos. We make use of anomaly priors and propose an anomaly-led sampling (AS) to enable that anomalous clips are more likely to be selected. Since frame-level annotations are unknown, it is impossible to directly identify anomalous clips. To solve this problem, we leverage a weakly supervised anomaly detector to predict clip-level anomaly confidences $\boldsymbol{l} \in \mathbb{R}^T$, where $l_i \in [0,1]$. With $\boldsymbol{l}$ in hands, we expect that for a clip, the probability of being selected is positively correlated with its anomaly confidence. A natural way is to select the top several clips with the highest anomaly confidence, but such a solution is too strict to be flexible. We believe that those clips with low anomaly confidences should also have a certain probability to be selected, on the one hand for data augmentation, on the other hand for salvaging false negatives of the anomaly detector. Taking inspiration from the selection strategy of evolutionary algorithms~\cite{wu2020evolutionary,jin2018data}, our anomaly-led sampling is based on the classical roulette-wheel selection~\cite{back1996evolutionary}. To be specific, we regard anomaly confidences $\boldsymbol{l}$ as the fitness, and then normalize all values to the interval [0,1] to ensure summation of selection probabilities equals one,

\begin{equation}
  p_i = \frac{\exp{(l_i/\tau)}}{\begin{matrix} \sum_{k=1}^T \exp{(l_k/ \tau)} \end{matrix}}
  \label{eq:prob}
\end{equation}
where $\boldsymbol{p}$ is selection probabilities, and $\tau$ is a temperature hyper-parameter~\cite{wu2018unsupervised}. Then calculate cumulative probabilities,
\begin{equation}
  q_i = \begin{matrix} \sum_{k=1}^i p_k \end{matrix}
  \label{eq:cum}
\end{equation}
It should be noted that $q_0=0$ and $q_T=1$. The final step, then, is to generate $N$ uniformly distributed random numbers in the interval [0, 1]. For each generated number $r$, the $i$-th feature in $\boldsymbol{F}(v)$ is selected if $q_{i-1}<r\le q_i$. A sequence with $N$ clip-level features is assembled in such a way, where the larger the anomaly confidence of a clip is, the more likely is its selection. We present the algorithm flow of Anomaly-led sampling in Algorithm~1.

\begin{figure}[!t]
 \removelatexerror
  \label{alg:roulette}
\begin{algorithm}[H]
\caption{\textbf{:} Anomaly-led sampling based on roulette-wheel selection}

\KwIn{anomaly confidence: $\boldsymbol{l}$; video features: $\boldsymbol{F}(v)$}
\KwOut{$N$ clip-level features}
step1: Compute the selection probability\;

$\quad \qquad p_i \gets \frac{\exp{(l_i/\tau)}}{\begin{matrix} \sum_{k=1}^T \exp{(l_k/ \tau)} \end{matrix}}$\

step2: Compute the cumulative probability\;

$\qquad \quad \qquad q_i \gets \begin{matrix} \sum_{k=1}^i p_k \end{matrix}$\;

$\qquad \quad \qquad q_0 \gets 0; q_T \gets 1$\

$k \gets 0$\;

\While{$k < N$}{
  Step3: Generate a random number $r \in [0,1]$\;
  \tcp{uniform distribution}
  
  step4: Select features\;
  
  \If{$ q_{i-1}< r \leq q_{i}$}{
    $i$-th feature in $\boldsymbol{F}(v)$ is selected;
  }
  $k \gets k + 1$\; 
}
\end{algorithm}
\end{figure}

This feature sequence based on anomaly-led sampling are mainly applied to cover anomalous segments, meanwhile, we also use fixed-frame sampling, e.g., uniform or random, to generate another sequence with $N$ clips for the entirety and normal scenarios.

\subsection{Video Prompt Based Masked Phrase Modeling}
\label{sec: mpm}

We propose a novel pretext task, i.e., video prompt based masked phrase modeling, for cross-modal fine-grained associations in video-text anomaly retrieval. VPMPM takes video representations and text representations as input and predicts the masked phrases, which is related to the prevalent masked language modeling in nature language processing. The main difference lies in that (1) VPMPM masks and predicts noun phrases and verb phrases instead of randomly selected words. Unlike single words, noun phrases and verb phrases comprise words of different parts of speech, e.g., nouns, adjectives, verbs, adverbs, etc., better correspond to the local objects and motions in video frames; (2) VPMPM fuses video representations with text representations through cross-modal attention, where video representations serve as fixed prompts~\cite{hou2022milan}. Such two specific designs encourage video encoder and text encoder to capture cross-modal and contextual representation interactions.

To achieve this pretext task, we introduce a Prompting Decoder, which is a standard decoder layer used in the Transformer. Since VPMPM involves the objectives of predicting masked noun phrases and masked verb phrases, Prompting Decoder needs to process noun phrases and verb phrases separately in a parameter-shared manner. Given the final video frame-level representations $X^v$ and text word-level $X^t$, we first randomly replace a noun phrase or verb phrase representations with mask embeddings~\cite{he2022masked}, where each mask token is a
shared, learned vector. Here we denote this masked text representation as $\widehat{X}^t$. Then we take $\widehat{X}^t$ as the query, and $X^v$ as the key and value, feed them into Prompting Decoder to predict the masked contents.

\subsection{Cross-Modal Alignment}
\label{sec: alignment}
In this paper, cross-modal alignment is used to match representations of different modalities, e.g., video-text and video-audio, from two complementary perspectives. Hence, we deal with CLS alignment and AVG alignment. Unless otherwise stated, here we take video-text as an example to describe these two alignments.

\noindent \textbf{CLS alignment.} CLS alignment is intended to compute the similarity between $g^v$ and $g^t$, and the similarity between them is a weighted sum~\cite{gabeur2020multi}, which is computed as, 
\begin{equation}
  s^g(v, t) = w_{ta} cos(\boldsymbol{g}^{vo}, \boldsymbol{g}^{to})+w_{tm} cos(\boldsymbol{g}^{vm}, \boldsymbol{g}^{tm}) 
  \label{eq:gsim}
\end{equation}
where $cos(\cdot,\cdot)$ is the cosine similarity between two vectors. $w_{ta}$ and $w_{tm}$ are weights, which are obtained from $\boldsymbol{g}^{ta}$ and $\boldsymbol{g}^{tm}$, respectively. Specifically, we pass $\boldsymbol{g}^{ta}$ ($\boldsymbol{g}^{tm}$) through a linear layer with softmax normalization, and output $w_{ta}$ ($w_{tm}$). 

\noindent \textbf{AVG alignment.} AVG alignment is intended to compute the similarity $s^h(v, t)$ between $h^v$ and $h^t$, which is same as CLS alignment. Notably, AVG alignment introduces more fine-grained information. The similarity is presented as,
\begin{equation}
  s^h(v, t) = w_{ta} cos(\boldsymbol{h}^{vo}, \boldsymbol{h}^{to})+w_{tm} cos(\boldsymbol{h}^{vm}, \boldsymbol{h}^{tm}) 
  \label{eq:hsim}
\end{equation}

\subsection{Training Objectives}
\label{sec: training objectives}
The final similarity between $v$ and $t$ is the weighted sum of $s^g(v,t)$ and $s^h(v,t)$, namely,
\begin{equation}
  s(v,t) = \alpha s^g(v,t) + (1-\alpha)s^h(v,t)
  \label{eq:sim}
\end{equation}
where $\alpha$ is a hyper-parameter, which lies in the range of [0,1]. Following the previous work~\cite{gabeur2020multi}, we obtain the bi-directional max-margin ranking loss, which is given by,
\begin{equation}
  \mathcal{L}_{align}=  \frac{1}{B}\sum_{i=1}^B \sum_{j\ne i} \left[\left[s_{ij}-s_{ii}+\Delta	\right]_+ + \left[s_{ji}-s_{ii}+\Delta \right]_+\right]
  \label{eq:loss0}
\end{equation}
where $B$ is batch size, $s_{ij}=s(v_i, t_i)$.

To optimize the weakly supervised anomaly detector in video encoder, we use the top-k strategy~\cite{paul2018w, wu2021learning} to obtain the video-level prediction from frame-level confidences $\boldsymbol{l}$, which is calculated as,
\begin{equation}
  \boldsymbol{l}=\frac{1}{k} \sum_{i=1}^{k} \boldsymbol{l}^{topk}_i
  \label{eq:topk}
\end{equation}
where $k=\lfloor \frac{T}{16} \rfloor$, and $\boldsymbol{l}^{topk}$ is the set of $k$-max frame-level confidences in $\boldsymbol{l}$ for the video $v$. We train this detector with binary cross-entropy loss $\mathcal{L}_{topk}$ between the video-level prediction $\rho^v$  and video-level binary label $y^v$,
\begin{equation}
  \mathcal{L}_{topk}= -\left[y^v log\left(\rho^v\right) + (1-y^v) log\left(1-\rho^v\right)\right]
 \label{eq:bce}
\end{equation}

For VPMPM in video-text anomaly retrieval, we adopt the cross-entropy loss $\mathcal{L}_{mpm}$ between the model’s predicted probability $\rho^t(\widehat{X}^t,X^v)$ 
and ground truth $y^{mask}$, which is presented as follows, 
\begin{equation}
  \mathcal{L}_{mpm}= -y^{mask} log\left(\rho^t(\widehat{X}^t,X^v)\right)
  \label{eq:mpm}
\end{equation}
where $y^{mask}$ is a one-hot vocabulary distribution.

At last, the overall loss is shown as follows,
\begin{equation}
  \mathcal{L}_{total} = \mathcal{L}_{align}+\lambda_1 \mathcal{L}_{topk}+\lambda_2 \mathcal{L}_{mpm}
  \label{eq:loss}
\end{equation}

\section{Experiments}
\subsection{Experimental Settings}
\noindent \textbf{Evaluation metrics.} Following prior works, we use the rank-based metric for performance evaluation, i.e., Recall at K (R@K, K=1, 5, 10), Median Rank (MdR), and Sum of all Recalls (SumR) to measure the overall performance.

\noindent \textbf{Implementation details.} We use Spacy\footnote{https://spacy.io/} to extract noun phrases and verb phrases. In video encoder and audio encoder, the anomaly detector is composed of 3 temporal convolution layers with kernel size of 7, the first layer has 128 units followed by 32 units and 1 unit layers. The first two layers are followed by ReLU, and the last layer is followed by Sigmoid. Dropout with rate of 0.6 is applied to the first two layers. In the text encoder, we use the ``BERT-base-cased model'' and fine-tune it with a dropout rate of 0.3. 

\noindent \textbf{Training.} We train our model with a batch size of 64 using Adam~\cite{kingma2014adam} optimizer. The initial learning rate is set as $5\times10^{-5}$ and decays by a multiplicative factor 0.95 per epoch. For hyper-parameters, hidden size $d$ is set as 768, and temperature parameter $\tau$ in Eq.~\ref{eq:prob} is set as 0.7. Empirically, we found the weight ratio $\alpha$=0.5 in Eq.~\ref{eq:sim} and sampling length $N$=50 worked well across different benchmarks. As the setup in~\cite{gabeur2020multi}, the margin $\Delta$ in Eq.~\ref{eq:loss0} is set as 0.05. $\lambda_1$ and $\lambda_2$ in Eq.~\ref{eq:loss} is set as 0.1 and 0.01, respectively, such a setup achieves optimal performance.

\begin{table*}[t]
  \centering
  \caption{Comparisons with the state-of-the-art methods on UCFCrime-AR.}
  \begin{tabular}{l|cccc|cccc|c}
    \toprule
    \multirow{2}{*}{Method}  & \multicolumn{4}{c|}{Text$\rightarrow$Video} & \multicolumn{4}{c|}{Video$\rightarrow$Text} & \multirow{2}{*}{SumR$\uparrow$}\\
    ~ & R@1$\uparrow$ & R@5$\uparrow$ & R@10$\uparrow$ & MdR$\downarrow$ & R@1$\uparrow$ & R@5$\uparrow$ & R@10$\uparrow$ & MdR$\downarrow$ \\
    \hline
    Random Baseline & 0.3 & 2.1 & 3.4 & 144.0 & 0.3 & 1.0 & 3.1 & 145.5 & 10.2 \\
    CE \cite{liu2019use} & 6.6 & 19.7 & 32.4 & 23.5 & 5.5 & 19.7 & 32.4 & 21.0 & 116.3 \\
    MMT \cite{gabeur2020multi} & 8.3 & 26.2 & 39.3 & 16.0 & 7.2 & 23.1 & 39.0 & 16.0 & 143.1 \\
    T2VLAD \cite{wang2021t2vlad} & 7.6 & 23.4 & 39.7 & 15.5 & 6.2 & \textbf{27.9} & 43.1 & 14.0 & 147.9 \\
    X-CLIP \cite{ma2022x} & 8.2 & 27.2 & 41.7 & 16.0 & 6.9 & 25.8 & 40.3 & 15.0 & 150.1 \\
    \hline
    HL-Net \cite{wu2020not}& 5.5 & 20.2 & 38.3 & 19.5 & 5.5 & 22.8 & 35.5 & 20.0 & 127.8 \\
    XML \cite{lei2020tvr} & 6.9 & 24.1 & 42.4 & 14.0 & 6.6 & 25.9 & 43.4 & 13.0 & 149.3 \\
    \hline
    \textbf{ALAN} & \textbf{9.0} & \textbf{27.9} & \textbf{44.8} & \textbf{14.0} & \textbf{7.3} & 24.8 & \textbf{46.9} & \textbf{12.0} & \textbf{160.7} \\
    \bottomrule
  \end{tabular}
   
   \label{tab:comparison_ucf}
\end{table*}

\begin{table*}[t]
  \centering
   \caption{Comparisons with the state-of-the-art methods on XDViolence-AR.}
  \begin{tabular}{l|cccc|cccc|c}
    \toprule
    \multirow{2}{*}{Method}  & \multicolumn{4}{c|}{Audio$\rightarrow$Video} & \multicolumn{4}{c|}{Video$\rightarrow$Audio} & \multirow{2}{*}{SumR$\uparrow$}\\
    ~ & R@1$\uparrow$ & R@5$\uparrow$ & R@10$\uparrow$ & MdR$\downarrow$ & R@1$\uparrow$ & R@5$\uparrow$ & R@10$\uparrow$ & MdR$\downarrow$ \\
    \hline
    Random Baseline & 0.4 & 0.6 & 2.5 & 399.5 & 0.1 & 0.6 & 0.8 & 399.5 & 5.0 \\
    CE \cite{liu2019use} & 11.4 & 33.3 & 47.0 & 12.5 & 13.0 & 34.3 & 46.4 & 13.0 & 185.4 \\
    MMT \cite{gabeur2020multi} & 20.5 & 53.5 & 68.0 & 5.0 & 23.0 & 54.6 & 69.5 & 5.0 & 289.1 \\
    T2VLAD \cite{wang2021t2vlad} & 22.4 & 56.1 & 71.0 & 4.0 & 23.2 & 57.1 & 73.5 & 4.0 & 303.3 \\
    X-CLIP \cite{ma2022x} & 26.4 & 61.1 & 73.9 & 3.0 & 26.4 & 61.3 & 73.8 & 4.0 & 322.9 \\
    \hline
    HL-Net \cite{wu2020not} & 12.4 & 36.6 & 48.3 & 11.0 & 13.4 & 38.3 & 52.1 & 10.0 & 201.1 \\
    XML \cite{lei2020tvr} & 22.9 & 55.6 & 70.3 & 5.0 & 22.6 & 57.4 & 71.4 & 4.0 & 300.2 \\
    \hline
    \textbf{ALAN} & \textbf{29.8} & \textbf{68.0} & \textbf{82.0} & \textbf{3.0} & \textbf{32.3} & \textbf{70.0} & \textbf{82.3} & \textbf{3.0} & \textbf{364.4} \\
    \bottomrule
  \end{tabular}
  
   \label{tab:comparison_xd}
\end{table*}

\subsection{Comparison with State-of-the-Art Methods}
We conduct experiments on UCFCrime-AR and XDViolence-AR and compare our ALAN with several recent methods that are widely used in video retrieval, video moment retrieval and VAD. 
CE~\cite{liu2019use}, MMT~\cite{gabeur2020multi}, T2VLAD~\cite{wang2021t2vlad}, and X-CLIP~\cite{ma2022x} are video retrieval methods; XML~\cite{lei2020tvr} is a video moment retrieval method, here it is used to retrieve videos, where the moment localization part is removed since moment annotations are unavailable in VAR. HL-Net~\cite{wu2020not} is a VAD method, since VAD is quite distinct from VAR, it is hard to directly use VAD method for VAR, here, we modify it as a video encoder for VAR. All methods use BERT to extract language features except CE that uses the word2vec word embeddings~\cite{mikolov2013efficient}.
We present comparison results in Tables~\ref{tab:comparison_ucf} and~\ref{tab:comparison_xd}, and observe that our ALAN shows a clear advantage over comparison methods in both text-video and audio-video VAR. Specifically, ALAN outperforms CE, MMT, T2VLAD, X-CLIP, HL-Net, and XML on UCFCrime-AR by 44.4, 17.6, 12.8, 10.6, 32.9, and 11.4 in terms of SumR, respectively. Furthermore, ALAN also achieves clear improvements against competitors on XDViolence-AR, which achieves a significant performance improvement of 41.5 in terms of SumR over the previous best method. Moreover, It can be found that, in comparison to the video and text, the video and audio are easier to align. We argue that video and audio are synchronous with concordant granularity, thereby leading to better align performance in VAR.

\subsection{Ablation Studies}
\noindent \textbf{Study on anomaly-led sampling.} As aforementioned, we propose a novel sampling mechanism, i.e., anomaly-led sampling, which combines with the ordinary fixed-frame sampling, and the joint effort is devoted to capturing local anomalous segments as well as overall information. To investigate the effectiveness of anomaly-led sampling, we conduct experiments on two benchmarks, and show results on Tables~\ref{tab:as-fs_ucf} and~\ref{tab:as-fs_xd}. As we can see from the first two rows, only using fixed-frame sampling or anomaly-led sampling results in a clear performance drop on both UCFCrime-AR and XDViolence-AR. {Besides, using anomaly-led sampling is inferior to using fixed-sampling on XDViolence-AR, we discover that the main reason for this problem is that the anomaly-led sampling mechanism is applied to both video and audio, resulting in key segments misalignment to some extent.}
Moreover, we also investigate the effect of sampling length. From the third row, we found that increasing the sampling length from 50 to 100 does not dramatically improve performance, and fixed-frame sampling still lags behind the combination of fixed-frame sampling and anomaly-led sampling, even though they both have the same sampling length at the moment. It also clearly demonstrates that the joint effect between anomaly-led sampling and fixed-frame sampling enables our model to capture key anomalous segments as well as holistic data information, {thus facilitating cross-modal alignment under local-anomaly and global-video perspectives. For example, in Figure~\ref{fig:sampling}, video frames that are selected by anomaly-led sampling are aligned with the key anomaly descriptions, e.g., two car collided violently, a man in black lay on the ground and shot. On another hand, these video frames selected by fixed-frame sampling are aligned with the complete descriptions.}

\begin{table}[t]
  \centering
   \caption{Comparisons of different samplings on UCFCrime-AR.}
  \begin{tabular}{l|cc|cc}
    \toprule
    \multirow{2}{*}{Sampling}  &\multicolumn{2}{c|}{Text$\rightarrow$Video} & \multicolumn{2}{c}{Video$\rightarrow$Text}\\
     & R@1$\uparrow$ & R@10$\uparrow$  & R@1$\uparrow$ & R@10$\uparrow$ \\
    \hline
    FS ($N$=50) & 6.6 & 35.5 & 4.8 & 42.4\\
    AS ($N$=50) & 7.9 & 37.6 & 5.5 & 41.7 \\
    FS ($N$=100) & 6.6 & 37.6 & 6.2 & 40.3\\
    FS+AS ($N$=50) & \textbf{9.0} & \textbf{44.8} & \textbf{7.3} & \textbf{46.9} \\
    \bottomrule
  \end{tabular}
  
   \label{tab:as-fs_ucf}
\end{table}

\begin{table}[t]
  \centering
   \caption{Comparisons of different samplings on XDViolence-AR.}
  \begin{tabular}{l|cc|cc}
    \toprule
    \multirow{2}{*}{Sampling}  &\multicolumn{2}{c|}{Text$\rightarrow$Video} & \multicolumn{2}{c}{Video$\rightarrow$Text}\\
     & R@1$\uparrow$ & R@10$\uparrow$  & R@1$\uparrow$ & R@10$\uparrow$ \\
    \hline
    FS ($N$=50) & 29.6 & 80.4 & 31.1 & 80.9 \\
    AS ($N$=50) & 26.9 & 78.6 & 27.4 & 78.9 \\
    FS ($N$=100) & 28.5 & 81.0 & 29.8 & 81.8 \\
    FS+AS ($N$=50) & \textbf{29.8} & \textbf{82.0} & \textbf{32.3} & \textbf{82.3}\\
    \bottomrule
  \end{tabular}
  
   \label{tab:as-fs_xd}
\end{table}

\begin{table}[t]
  \centering
  \caption{VPMPM Studies on UCFCrime-AR.}
  \scalebox{0.95}{
  \begin{tabular}{l|cc|cc}
    \toprule
    \multirow{2}{*}{Method}  &\multicolumn{2}{c|}{Text$\rightarrow$Video} & \multicolumn{2}{c}{Video$\rightarrow$Text}\\
     & R@1$\uparrow$ & R@10$\uparrow$  & R@1$\uparrow$ & R@10$\uparrow$ \\
    \hline
     w/o VPMPM & 7.9 & 43.4 & 7.2 & 43.4 \\
     random words & 8.6 & 43.8 & 6.2 & 44.5 \\
     noun\&verb words & \textbf{10.0} & 44.5 & 6.6 & 42.8 \\
     noun\&verb phrases & 9.0 & \textbf{44.8} & \textbf{7.3} & \textbf{46.9} \\
    \bottomrule
  \end{tabular}}
  
   \label{tab:mpm}
\end{table}

\begin{table}[t]
  \centering
   \caption{Comparisons of different alignments on UCFCrime-AR.}
  \begin{tabular}{l|cc|cc}
    \toprule
    \multirow{2}{*}{Alignment}  &\multicolumn{2}{c|}{Text$\rightarrow$Video} & \multicolumn{2}{c}{Video$\rightarrow$Text}\\
     & R@1$\uparrow$ & R@10$\uparrow$  & R@1$\uparrow$ & R@10$\uparrow$ \\
    \hline
     CLS & 6.2 & 42.8 & \textbf{7.6} & 40.3 \\
     AVG & 6.6 & 33.8 & 4.8 & 36.9  \\
     CLS+AVG & \textbf{9.0} & \textbf{44.8} & 7.3 & \textbf{46.9} \\
    \bottomrule
  \end{tabular}
 
   \label{tab:align_ucf}
\end{table}

\begin{table}[!tb]
  \centering
  \caption{Comparisons of different alignments on XDViolence-AR.}
  \begin{tabular}{l|cc|cc}
    \toprule
    \multirow{2}{*}{Alignment}  &\multicolumn{2}{c|}{Audio$\rightarrow$Video} & \multicolumn{2}{c}{Video$\rightarrow$Audio}\\
     & R@1$\uparrow$ & R@10$\uparrow$  & R@1$\uparrow$ & R@10$\uparrow$ \\
    \hline
     CLS & 26.8 & 77.9 & 28.6 & 77.4 \\
     AVG & 28.0 & 79.3 & 30.0 & 80.1  \\
     CLS+AVG & \textbf{29.8} & \textbf{82.0}  & \textbf{32.3} & \textbf{82.3} \\
    \bottomrule
  \end{tabular}
   \label{tab:align_xd}
\end{table}

\begin{table}[t]
  \centering
  \caption{Detailed influences of $\alpha$ on UCFCrime-AR.}
  \begin{tabular}{c|cc|cc}
    \toprule
    \multirow{2}{*}{Value of $\alpha$}  &\multicolumn{2}{c|}{Audio$\rightarrow$Video} & \multicolumn{2}{c}{Video$\rightarrow$Audio}\\
     & R@1$\uparrow$ & R@10$\uparrow$  & R@1$\uparrow$ & R@10$\uparrow$ \\
    \hline
    0.0 & 6.6 & 33.8 & 4.8 & 36.9 \\
    0.2 & 8.3 & 38.6 & 6.2 & 40.3 \\
    \rowcolor{gray! 20}0.4 & 7.9 & 42.4 & 6.9 & 44.5 \\
    \rowcolor{gray! 20}0.5 & \textbf{9.0} & 44.8 & 7.3 & 46.9 \\
    \rowcolor{gray! 20}0.6 & 7.6 & \textbf{45.2} & 6.2 & \textbf{47.2} \\
    0.8 & 7.6 & 43.8 & 5.5 & 43.4 \\
    1.0 & 6.2 & 42.8 & \textbf{7.6} & 40.3 \\
    \bottomrule
  \end{tabular}
   \label{tab:alpha}
\end{table}

\begin{table}[t]
  \centering
  \caption{Detailed influences of $\alpha$ on UCFCrime-AR.}
  \begin{tabular}{c|cc|cc}
    \toprule
    \multirow{2}{*}{Value of $\alpha$}  &\multicolumn{2}{c|}{Audio$\rightarrow$Video} & \multicolumn{2}{c}{Video$\rightarrow$Audio}\\
     & R@1$\uparrow$ & R@10$\uparrow$  & R@1$\uparrow$ & R@10$\uparrow$ \\
    \hline
    0.0 & 28.0 & 79.3 & 30.0 & 80.1 \\
    0.2 & 30.4 & 80.8 & 32.1 & \textbf{82.6} \\
    \rowcolor{gray! 20}0.4 & \textbf{31.9} & 80.9 & 32.3 & 82.4 \\
    \rowcolor{gray! 20}0.5 & 29.8 & 82.0 & 32.3 & 82.3 \\
    \rowcolor{gray! 20}0.6 & 31.0 & \textbf{82.6} & \textbf{32.8} & 80.8 \\
    0.8 & 28.0 & 79.8 & 30.1 & 79.3 \\
    1.0 & 26.8 & 77.9 & 28.6 & 77.4 \\
    \bottomrule
  \end{tabular}
   \label{tab:alpha2}
\end{table}

\noindent \textbf{Study on VPMPM.} Here we conduct experiments to certify the advantage of VPMPM for video-text fine-grained associations. When ALAN removes VPMPM at training time, we observe the performance clearly drops as shown in Table~\ref{tab:mpm}. Besides, masking and predicting in the form of random words rather than noun phrases and verb phrases in VPMPM hurts performance. We can also see that, using noun phrases and verb phrases are superior to noun and verb words on most evaluation metrics. This demonstrates that noun phrases and verb phrases, as the sequences of words with different parts of speech, can better align with related local contents in videos.
\begin{figure}[!tb]
  \centering
   \includegraphics[width=1.0\linewidth]{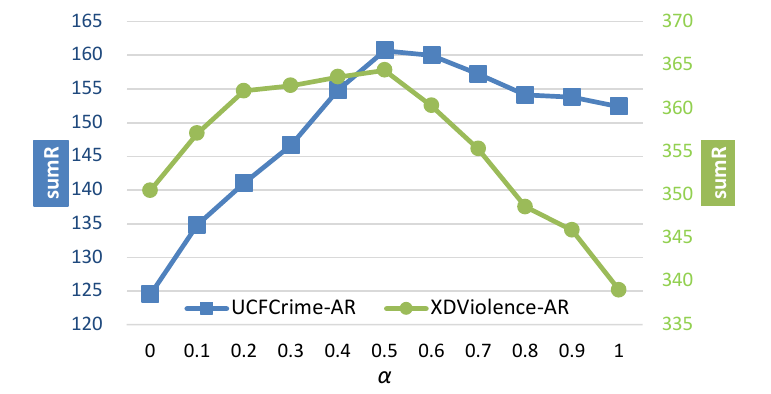}
   \caption{Influences of $\alpha$ on both UCFCrime-AR and XDViolence-AR.}
   \label{fig:alpha}
\end{figure}

\begin{figure*}[t]
  \centering
  \includegraphics[width=1.0\linewidth]{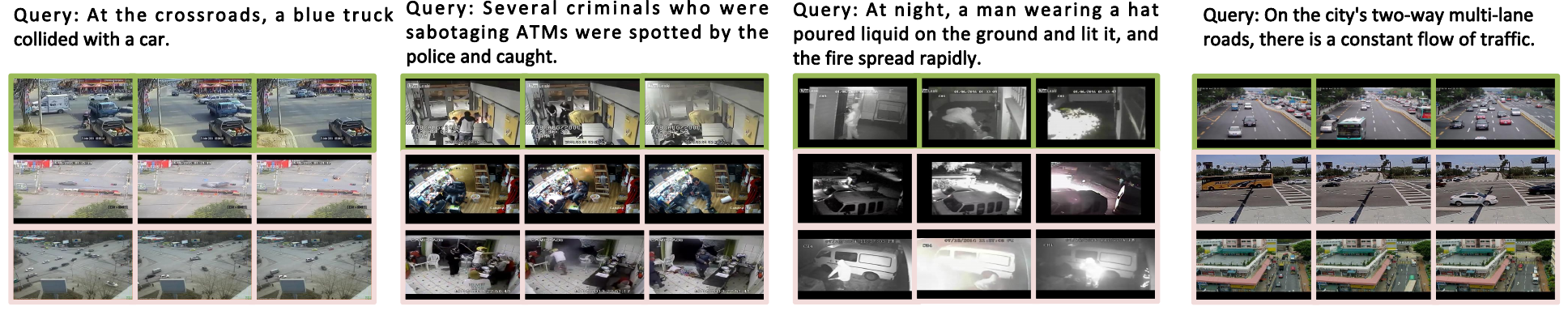}
  \caption{Some retrieval examples on UCFCrime-AR. We visualize top 3 retrieved videos (green: correct; pink: incorrect).}
  \label{fig:retrieval}
\end{figure*}

\noindent \textbf{Study on cross-modal alignment.} Tables~\ref{tab:align_ucf} and~\ref{tab:align_xd} present the performance of two different alignments in our ALAN. We found that CLS alignment and AVG alignment obtain worse results when used alone in comparison to the model of jointly using both. Such results demonstrate the complementarity of these two alignments. A key observation is the AVG alignment performs better than CLS alignment on XDViolence-AR, but the opposite is true on UCFCrime-AR, we suspect that video and audio are easier to align at the fine-grained level due to their concordant granularity. Moreover, we also investigate the influence of $\alpha$. We try $\alpha$ with its value ranging from 0.0 to 1.0 with an interval of 0.1. As shown in Figure~\ref{fig:alpha}, with the increase of $\alpha$, the performance gradually improves and then decreases, when $\alpha$ is set as 0.5, our method achieves the best performance. {In order to further explore how to choose $\alpha$, we also show the detailed retrieval results of different $\alpha$ in Tables~\ref{tab:alpha} and~\ref{tab:alpha2}. It is not hard to see that it is a balanced choice to set the value range of $\alpha$ to 0.4-0.6, where two different cross-modal alignments make nearly the same contribution.}

\subsection{Qualitative Analyses}
\noindent\textbf{Visualization of retrieval results.} Some text-to-video retrieval examples on UCFCrime-AR are exhibited in Figure~\ref{fig:retrieval}, where retrieval results of a normal video is shown at the far right. 
We observe ALAN successfully retrieves the related video given a text query, and there are considerable similarities between the top 3 retrieved videos. This also demonstrates VAR is a challenging task as some scenes are similar with delicate differences.

\begin{figure}[!tb]
  \centering
   \includegraphics[width=1.0\linewidth]{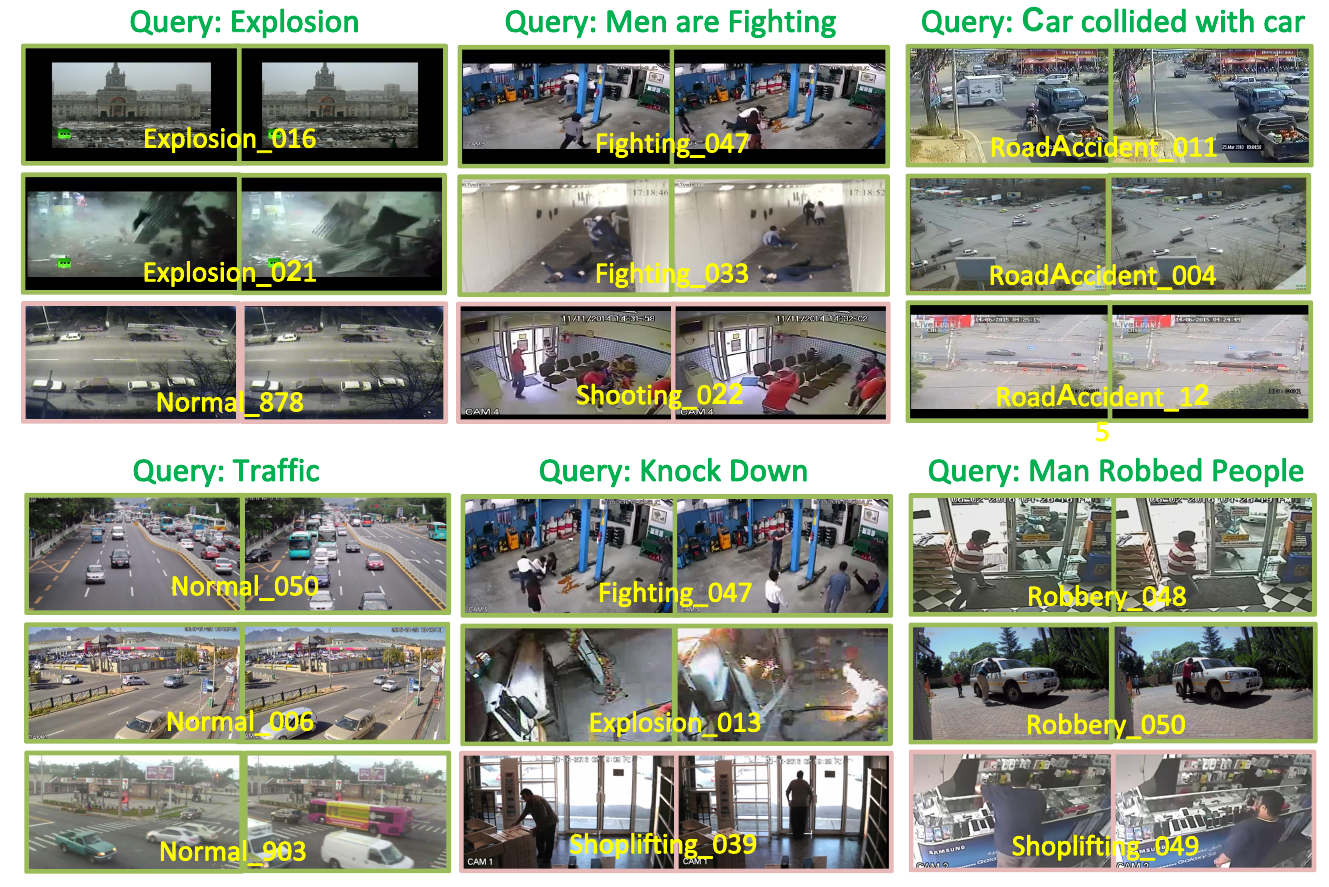}
   \caption{Some coarse caption retrieval examples on UCFCrime-AR.}
   \label{fig:coarse}
\end{figure}

\noindent\textbf{Visualization of coarse caption retrieval.} In VAR task, the purpose of using accurate captions is to distinguish fine differences and avoid being into a \textit{one-to-many} dilemma~\cite{wray2021semantic}. 
To further verify the generalization capacity of ALAN, we use several coarse captions that are not directly applied in model training to retrieve videos, results in Figure~\ref{fig:coarse} clearly show that ALAN works very well with different lengths of coarse captions, and also demonstrate ALAN has learned several abstract semantic information, e.g., explosion, fighting, traffic. {This also convincingly indicates our methods can meet practical requirements where users cannot provide a complete text description of the videos they intend to search, such as the example in the lower right of Figure~\ref{fig:coarse}, users give the the retrieval model a incomplete description ``man robbed people'', and the model returns top 3 related videos, in which the contents correspond to robbery, steal, man, and people.}

\begin{figure*}[!tb]
  \centering
  \includegraphics[width=1.0\linewidth]{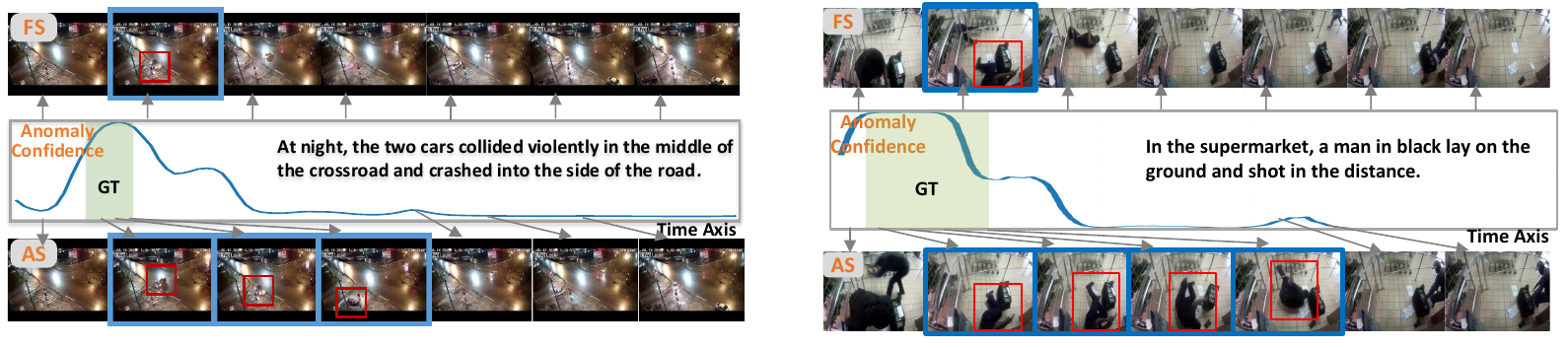}
  \caption{Different samplings for video frame selection. Left: road accident; Right: Shooting.}
  \label{fig:sampling}
\end{figure*}
\begin{figure*}[t]
  \centering
  \includegraphics[width=1.0\linewidth]{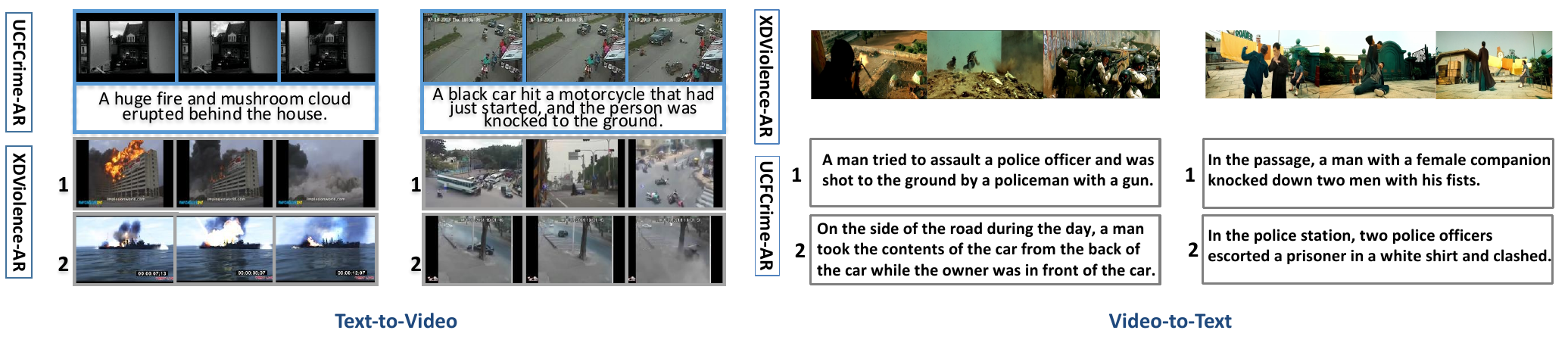}
  \caption{Zero-shot retrieval results. The left two columns present zero-shot text-to-video anomaly retrieval, and the right two columns present zero-shot video-to-text anomaly retrieval. }
  \label{fig:zeroshot}
\end{figure*}

\noindent\textbf{Visualization of anomaly-led sampling.} We visualize video frames selected by fixed-frame sampling and anomaly-led sampling in Figure~\ref{fig:sampling}. These examples are taken from videos of road accident and shooting scenes. It can be seen from the second row that the duration of anomalous event accounts for less than one-fifth of the entire video length, therefore, frames related to the anomalous event are hard to select based on fixed-frame sampling. In stark contrast to fixed-frame sampling, anomaly-led sampling is based on anomaly confidences generated by the anomaly detector, and it can select more frames related to anomalous events since the probability of being selected has positive correlations with anomaly confidences, where anomaly detector generates high confidences in anomalous segments which is shown in the second row.

\noindent\textbf{Visualization of zero-shot retrieval.} ALAN is trained on UCFCrime-AR and XDViolence-AR for text-video and audio-video anomaly retrieval, respectively. Moreover, scenarios in these two benchmarks are different, because videos from UCFCrime-AR are captured with fixed cameras, whereas videos from XDViolence-AR are collected from movies and YouTube. Here we explore that, given a cross-modal query from UCFCrime-AR (or XDViolence-AR), is ALAN trained on UCFCrime-AR (or XDViolece-AR) capable of retrieving some relevant videos from XDViolence-AR (or texts from UCFCrime-AR)? We show the top 2 retrieval results in Figure~\ref{fig:zeroshot}.
In text-to-video anomaly retrieval, we found that given text queries from UCFCrime-AR, ALAN can retrieve some videos from XDViolence-AR that look semantically plausible, even if there are no completely relevant videos in XDViolence-AR. Interestingly, the video in the bottom left is an animation. ALAN learns several local semantic contents and retrieves videos based on these local semantic contents, such as ``huge fire'' and ``mushroom cloud''. 
In video-to-text anomaly retrieval, although retrieved text descriptions are not completely related, ALAN captures partial semantic information from movie videos, such as ``a man'', ``a female companion'', ``knock down somebody with fists'', etc.

\subsection{Running Time}
We report the retrieval time for UCFCrime-AR with 290 video-text test pairs and XDViolence-AR with 800 video-audio test pairs, our method costs 2.7s and 5.6s, respectively. Generally, it only needs about 0.008s to process a pair on both datasets, showing its higher efficiency. {The reason why our method remains high retrieval efficiency is that it has a dual-encoder structure during the test stage, that is, using two separate encoders to embed video and text features and project them into the latent joint space, and only the cosine similarity between video and text features is calculated as similarity, without complicated and inefficient cross-modal interactions. However, it is worth noting that, during the training phase, our method integrates text and video as inputs to a joint encoder for the cross-modality fusion, which can establish local correlation between video-text features and improve retrieval accuracy. Therefore, our method obtains the advantages of the above two kinds of methods, that is, achieving fine-grained video-text interactions while maintaining high retrieval efficiency.}

\section{Conclusion}
In this paper, we introduce a new task called video anomaly retrieval to remedy the inadequacy of video anomaly detection in terms of abnormal event depict, further facilitate video anomaly analysis research in cross-modal scenarios. We construct two VAR benchmarks, i.e., UCFCrime-AR and XDViolence-AR, based on popular VAD datasets. Moreover, we propose ALAN which includes several components, where anomaly-led sampling is used to capture local anomalous segments, which coordinates with ordinary fixed-frame sampling to achieve complementary effects; Video prompt based masked phrase modeling is used to learn cross-modal fine-grained associations; Cross-modal alignment is used to match cross-modal representations from two perspectives. The future work will lie in two aspects, 1) exploiting cross-modal pre-trained models to capture more powerful knowledge for VAR; 2) leveraging VAR to assist VAD methods for more precise anomaly detection.

\bibliographystyle{IEEEtran}
\bibliography{ref}

\end{document}